\documentclass[journal]{IEEEtran}

\usepackage{times}
\usepackage{graphicx} 
\usepackage{subfigure} 

\usepackage{algorithm}
\usepackage{algorithmic}

\usepackage{helvet}
\usepackage{courier}

\usepackage{makecell}
\usepackage{graphicx}
\usepackage{subfigure}
\usepackage{color}
\usepackage{amsfonts}
\usepackage{amsmath}
\usepackage{amssymb}

\usepackage{multirow}
\usepackage{booktabs}

\def\etal{{\em et al.\/}\, }

\DeclareMathAlphabet\mathbfcal{OMS}{cmsy}{b}{n}

\def\0{{\bf 0}}
\def\1{{\bf 1}}

\def\bx{{\bf x}}
\def\by{{\bf y}}
\def\bz{{\bf z}}

\def\mmR{{\mathbb R}}

\def\bx{{\bf x}}

\def\by{{\bf y}}

\def\bz{{\bf z}}

\usepackage{ntheorem}

\newtheorem*{*thm}{Theorem}

\newtheorem*{*lemma}{Lemma}

\usepackage{subfigure}

\begin{document}
	
	\title{Auto-Embedding Generative Adversarial Networks for High Resolution Image Synthesis}
	
	\author{
		Yong Guo$^*$\thanks{$^*$ Authors contributed equally.},
		Qi Chen$^*$,
		Jian Chen$^*$,
		Qingyao Wu,
		Qinfeng Shi,
		and Mingkui Tan$^\dag$\thanks{$^\dag$ Corresponding author.}
	}
	
	\maketitle
	
	\begin{abstract}
		Generating images via generative adversarial network (GAN) {has} attracted much attention recently. However, most of the existing {GAN-based} methods can only produce low resolution images of limited quality.
		Directly generating high resolution images using GANs is nontrivial, and often produces problematic images with incomplete objects.
		{To address this issue,} we develop a novel GAN called Auto-Embedding Generative Adversarial Network (AEGAN), which {simultaneously} encodes the global structure features and captures the 
		fine-grained details. In our network, we use an autoencoder to learn the intrinsic high-level structure of real images and design a novel denoiser network to 
		{provide photo-realistic details for the generated images.}
		In the experiments, we are able to produce $512 \times 512$ images of promising quality directly from the input noise. 
		The resultant images exhibit better perceptual photo-realism, {{\it i.e.,} with sharper structure and richer details}, than other baselines on several datasets, including Oxford-102 Flowers, Caltech-UCSD Birds (CUB), High-Quality Large-scale CelebFaces Attributes (CelebA-HQ), Large-scale Scene Understanding (LSUN) and ImageNet.
	\end{abstract}
	

	\IEEEpeerreviewmaketitle
	
	\section{Introduction}
	
	\IEEEPARstart{B}{uilding} a generative model {that} produces photo-realistic images of high resolution has been a challenging problem in the field of computer vision. 
	{Compared to low resolution images,} the high resolution images of promising quality often provide richer information and benefit the training of deep neural networks (DNNs) in many real-world applications,
	such as texture synthesis~\cite{jetchev2016texture,ulyanov2016texture, ndjiki2011depth, sun2016hems}, super-resolution~\cite{ledig2016photo, shi2016real, yang2013self} and {attribute editing~\cite{he2017arbitrary, yu2018super, lee2017photo}.}
	However, how to produce high-quality data when increasing the image resolution still remains an open question.
	
	Recently, generative adversarial networks (GANs)~\cite{goodfellow2014generative} have achieved great success and become the workhorse of many challenging tasks, including
	image generation~\cite{radford2015unsupervised,arjovsky2017wasserstein,cao2018adversarial}
	video prediction~\cite{ranzato2014video,mathieu2015deep} and image translation~\cite{radford2015unsupervised,isola2016image,zhu2017unpaired,yi2017dualgan}. Typically, GANs learn to generate data by playing a two-player game: a generator {attempts} to produce samples from a simple prior distribution {({\it e.g.}, Gaussian distribution),} while a discriminator acts as a judge to distinguish the generated data from the real one.
	
\begin{figure}[t]
		\centering
		\includegraphics[width=1\columnwidth]{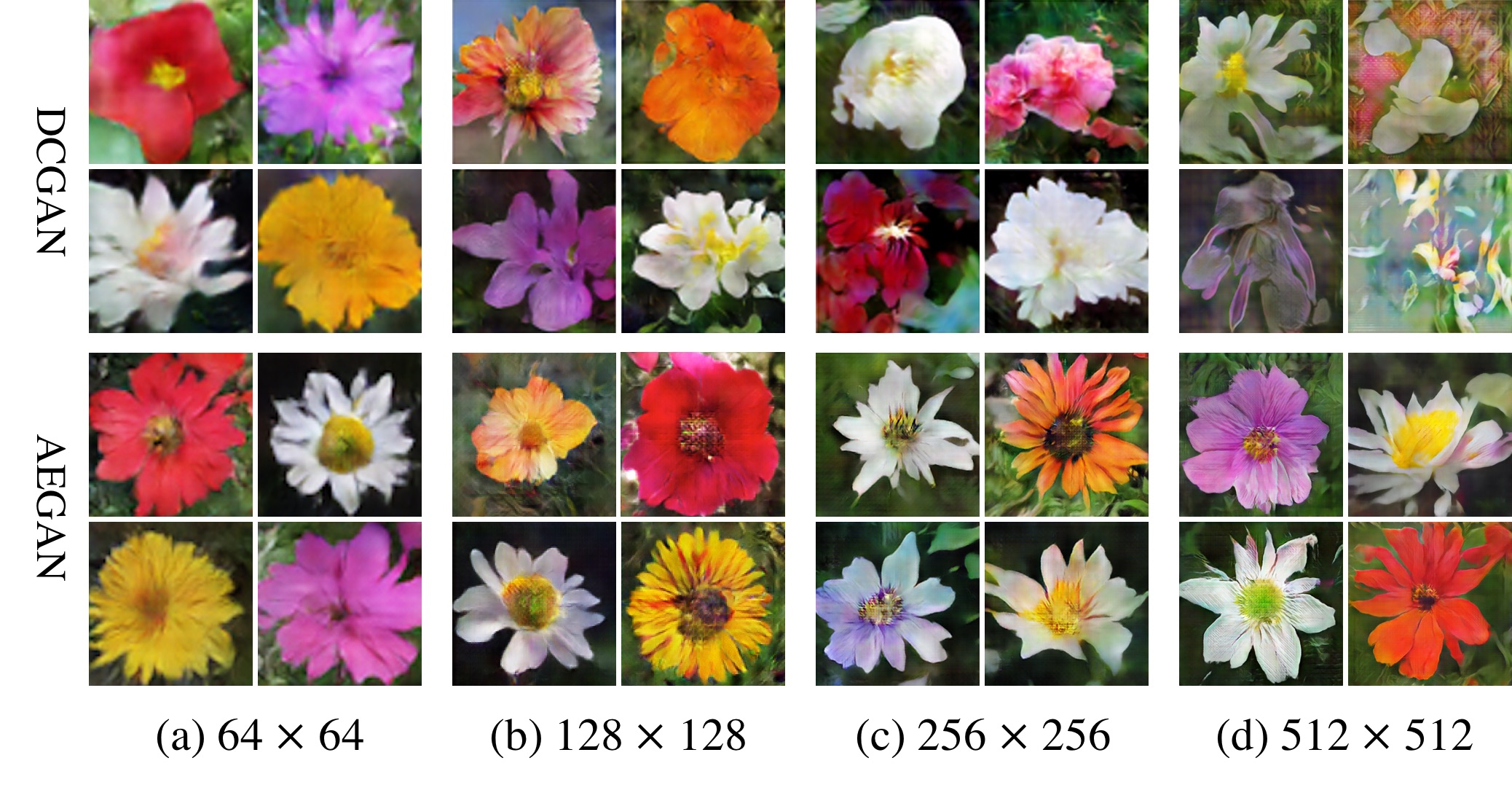}
		\caption{Samples generated by DCGAN (top) and AEGAN (bottom) of different resolutions on the Oxford-102 Flowers dataset.}
		\label{fig:varying-resolution}
	\end{figure}

	When generating images directly from the prior distribution, 
	the quality of {the images generated}
	by most of the existing models can be quite limited, especially when synthesizing very high resolution images. To be specific, deep generative models often produce meaningless images {that} may contain multiple distorted regions with {blurred} structure~\cite{odena2016conditional}. To illustrate this issue, in Fig.~\ref{fig:varying-resolution}, 
	we compare generated samples of different resolutions produced by the well-known model DCGAN~\cite{radford2015unsupervised} and {by} our proposed method. 
	From Fig.~\ref{fig:varying-resolution} top, the low resolution images of size $64 \times 64$ and $128 \times 128$ preserve clear and complete object structure, including stamen, petals, etc. However, for the {higher resolutions} of $256 \times 256$ or $512 \times 512$, DCGAN fails to produce meaningful samples and yields very poor results {compared to low resolutions}. 
	{Regarding this issue, it is necessary and important to explore a new method to improve the performance of generative models when increasing the image resolution.}
	
	The difficulties of generating high resolution images are generally attributed to two reasons. 
	{The first is that} it is hard to directly learn a mapping between the prior distribution (\textit{i.e.}, input noise) 
	and the distribution of high-dimensional real data, due to {the} large distribution gap.
	Since high-dimensional data often lie on some low-dimensional manifold~\cite{tenenbaum2000global, roweis2000nonlinear, chen2018low},
	we use low-dimensional embedding to uncover the {image's}
	structural information, which acts as a bridge to connect the prior distribution with the distribution of real data. This embedding is extracted by an autoencoder {that is able to reconstruct} the real images while preserving clear photo-structure.
	More critically, matching a low-dimensional embedding representation can significantly ease the training of GANs compared to learning high resolution images directly.
	
	The second reason regarding this issue is that there is no additional knowledge, such as label or semantic information obtained from real data, to help the model training.
	Note that the prior distribution is usually very simple, e.g., standard Gaussian distributions or {uniform} distributions~\cite{goodfellow2014generative}, and {is} independent of {the} real data, which may lose the original semantic information.
	In practice, GANs can easily lose the primary characteristics of data 
	{that are used by humans in image recognition,}
	resulting in meaningless images (see results in Section~\ref{sec:exp}). 
	On the contrary, the extracted embedding, which contains the spatial structural information, is able to act as a kind of additional knowledge to improve the training.
	
	By translating the input noise into an embedding and decoding it into a corresponding synthetic image, however,
	the model may introduce many noisy artifacts.
	To alleviate this issue, we further develop an adversarial denoiser network to enhance the photo-realism of the generated images. This denoiser model takes synthetic images as input and forces them to be perceptually indistinguishable from the real images in terms of texture details. In practice, we feed the generated images into the denoiser model to remove these artifacts and provide photo-realistic details.
	
	Based on the above intuitions, we propose the Auto-Embedding Generative Adversarial Networks (\textbf{AEGANs}) for high resolution image synthesis. 
	{Unlike the considered baselines, AEGAN is able to consistently generate perceptually promising images at different resolutions (see Fig.~\ref{fig:varying-resolution} bottom).}
	Moreover, the proposed method significantly outperforms the alternative approaches and produces photo-realistic $512 \times 512$ images, which simultaneously capture the high-level photo-structure and preserve the low-level details.
	
	In this paper, we make the following contributions:
	\begin{itemize}
		\item We devise a novel Auto-Embedding Generative Adversarial Net (AEGAN) 
		{that generates}
		high resolution images by learning a latent embedding extracted from an autoencoder. 
		{AEGAN exploits the high-level photo-structure and acts as a bridge to connect the distributions of the input noise and real data.} 
		As a result, the proposed method can produce much more meaningful images with clear structure and rich details.
		\item We develop a new denoiser network {that uses}
		an encoder-decoder network as the generator to remove artifacts and provide 
		photo-realistic details in the generated images.
		\item The proposed method is able to produce high resolution images with promising quality. For example, when the desired resolution is $512 \times 512$, the generated images are of much higher quality than those obtained {using} the considered methods.
	\end{itemize}
	
	\section{Related Work}
	
	\subsection{Generative Adversarial Networks}
	{Recently, Generative Adversarial Networks (GANs)~\cite{goodfellow2014generative} have shown promising performance for generating natural images,
		such as DCGAN~\cite{radford2015unsupervised}, WGAN~\cite{arjovsky2017wasserstein}, WGAN-GP~\cite{gulrajani2017improved}, LAPGAN~\cite{denton2015deep}, {\it etc}.
		More recently, Chen~\etal propose a modification of generators in GANs, called Self-Modulation framework~\cite{chen2018self} {that} improves the performance of {the} generated images 
		by modulating the architectural features of GAN generators.
		Moreover, GANs have also been applied to range of other interesting applications, such as text to image synthesis~\cite{reed2016generative, zhang2016stackgan, Han17stackgan2, Tao18attngan}, super-resolution~\cite{ledig2017photo, wang2018esrgan}, image inpainting~\cite{dolhansky2018eye, yang2017high, yu2018generative} and so on.
	}

	\subsection{High-Resolution Image Generation}
	{Generating high resolution images has gained much attention in the last few years in light of the advances in deep learning.
		To achieve this, one can compare the distribution divergences of the real and the generated data in a low-dimensional data space, such as StackGAN~\cite{zhang2016stackgan}, StackGAN-v2~\cite{Han17stackgan2}, AGE~\cite{ulyanov2017adversarial} and AttnGAN~\cite{Tao18attngan}.
	}
	
	{
		In StackGAN, the low resolution image obtained from Stage-I and the conditional text are used to produce high resolution image through Stage-II GAN, just {as is done in} super resolution. 
		{
			However, for unconditional image generation, the conditional text is not necessary and can be discarded. Moreover, the Stage-II GAN is trained by minimizing the JS distance between real and fake data. Unlike StackGANs,
			{in addition to minimizing the JS divergence,}
			we seek to increase the fidelity of images while preserving the image content by introducing a pixel-wise loss.
		}
	}
	{
		StackGAN-v2 is {an} extension of StackGAN {that} uses {a} cascade mechanism to generate the image in gradually growing resolution. Based on StackGAN-v2, AttnGAN introduces the attention mechanism to improve the visual quality of the generated images.
		AGE uses an autoencoder to project both the real data and the generated data into a latent space. To compare the distribution divergence, AGE aligns a simple prior distribution in the latent space and the data distribution in {the} image space.
		More recently, Zhang~\etal design a SAGAN model~\cite{zhang2018self} {that} incorporates the self-attention mechanism into the generator and discriminator. Based {SAGAN model,}
		Brock~\etal propose a BigGAN model~\cite{brock2018large} {that improves} the performance using a suite of tricks for training.}
	
	\subsection{Autoencoder GAN}
	{Autoencoders have been widely used in GANs and a plethora of attempts have been made to improve the training. For example, {the} adversarial autoencoder (AAE) leverages an autoencoder to match the distribution of latent embedding with the prior.
		Boundary-equilibrium GANs~\cite{berthelot2017began} and energy-based GANs~\cite{zhao2016energy} use an autoencoder as the discriminator to stabilize the training. Warde-Farley {\it et al.}~\cite{warde2016improving} combine an autoencoder loss with the GAN loss by matching high-level feature similarity. 
		CycleGANs~\cite{zhu2017unpaired} and DualGANs~\cite{yi2017dualgan}
		construct a bidirectional loss with autoencoder and GAN for data translation between two domains.
		{The methods described in~\cite{larsen2015autoencoding,rosca2017variational}, which are based on autoencoders, combine a Variational Autoencoder (VAE) with GAN}
		using variational inference to solve the intractability of the marginal likelihood in GAN.
		Plug and play generative networks (PPGN)~\cite{nguyen2016plug} combines an autoencoder loss with a GAN loss and a classification loss. 
	}
	
	{
		Very recently, Karras~\etal \cite{karras2017progressive} have proposed a new progressively-growing training method {that} gradually adds one more block per stage in both the generator and discriminator networks. In this way, it can significantly improve the synthetic image quality.
		However, this training method {takes {a} long {time} to converge} ({\it i.e.,} more than two days), while DCGANs {and}
		our proposed AEGANs only {require} several hours for training.
		Unlike other methods, in this paper, we use an autoencoder to transform high-dimensional data to a low-dimensional embedding in the latent space. The extracted embedding contains rich structural information of real images and aids the training to produce images of promising quality.}
	
	\begin{figure*}[t]
		\centering
		\includegraphics[width=2\columnwidth]{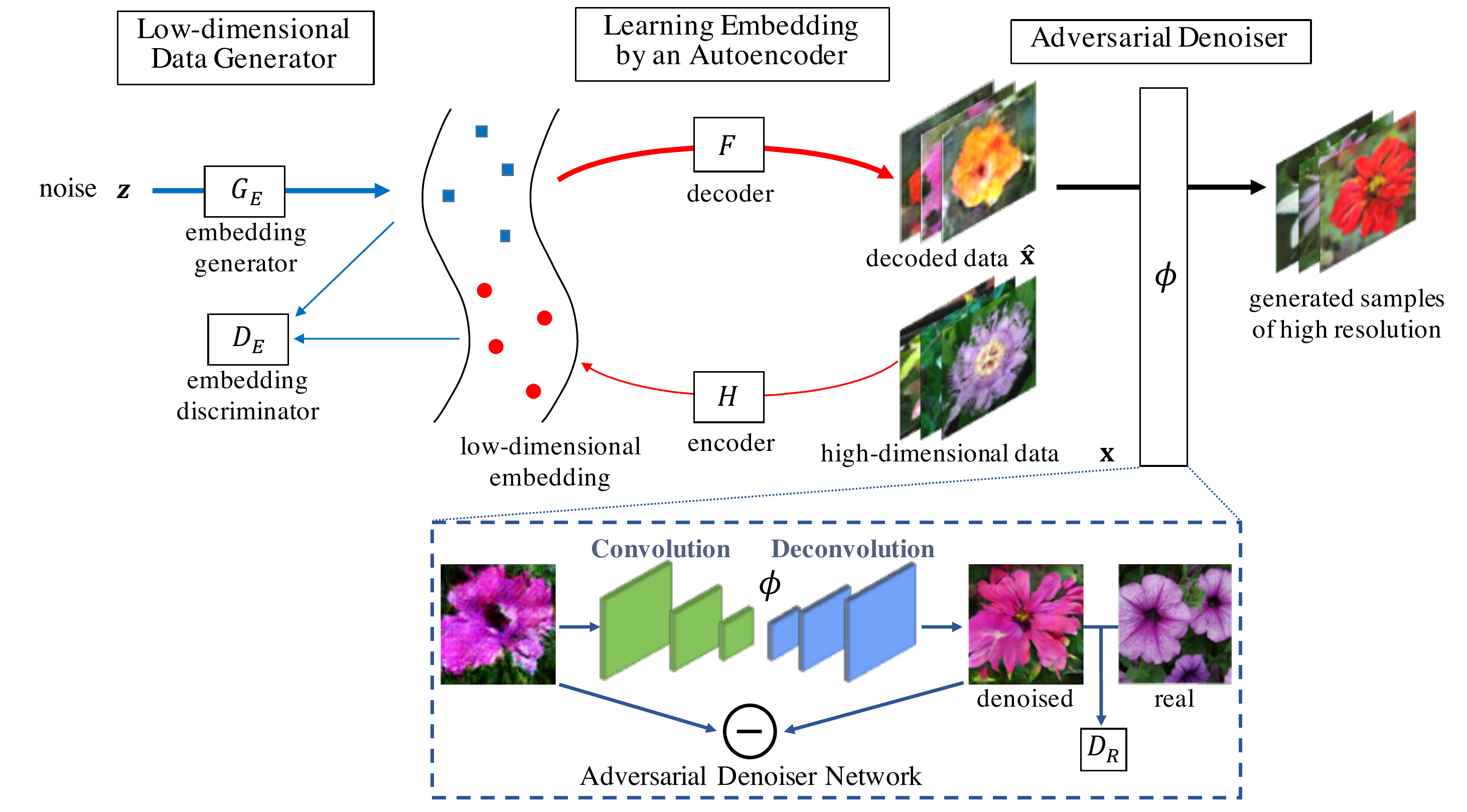}
		\caption{{Overall} architecture of the proposed method. An autoencoder is trained to encode the low-dimensional embedding of high-dimensional images. The embedding will be used to guide the learning of the generative model from the input noise. A denoiser network is devised to  enrich the details and denoise the artifacts in the generated images. Blue squares denote the synthetic embedding and the red circles denote the extracted embedding of real data. The bold lines in the figure indicate the main data stream to produce high resolution images.}
		\label{fig:network}
	\end{figure*}
	
	\section{The Proposed Method}
	
	{
		Considering that high-dimensional data often lie on some low-dimensional manifold, we seek to exploit a low-dimensional embedding extracted from large training data to improve the performance of GANs.
	}
	{To this end,}
	we present a novel generative method called Auto-Embedding Generative Adversarial Network (\textbf{AEGAN}). 
	Corresponding to the overall structure in Fig.~\ref{fig:network}, the detailed algorithm is shown in Algorithm~\ref{alg:training}.
	
	We propose to {train GANs by} matching the distribution of the low-dimensional embeddings on a latent manifold (see Step 1) instead of the distributions of pixel-level images, 
	{
		which is very different from the existing GANs methods.
	}
	Specifically, we use the embeddings extracted from an autoencoder to bridge the distribution gap between the input noise and real data. 
	{The embeddings often contain rich structural information and help to generalize well with high resolution images.}
	After using the autoencoder to extract the low-dimensional embedding, we learn a GAN in the embedding space to effectively exploit the structural information. 
	Moreover, we devise a denoiser network to remove artifacts/noises and refine the photo-realistic details (see Step 2).
	During training, we can use a fine-tuning strategy (which, however, is optional) to further improve the performance. 
	As shown in Fig.~\ref{fig:network}, the whole scheme mainly includes three important components, namely an autoencoder, a generative module, and a denoiser network.
	In this section, we will depict each part of the method in the following subsections.
	
	\subsection{Learning Embedding by Autoencoder}
	In this paper, we seek to bridge the distribution gap between the input noise and the real images using a latent embedding extracted from an autoencoder.
	The autoencoder contains an encoder $H$ which maps the high resolution images into a low-dimensional embedding and a decoder $F$ which translates the latent embedding back to high resolution images. Here, the encoder is a fully convolutional network {that extracts} the high-level {features} of {the} data, while the decoder is a fully deconvolutional model {that} effectively {recovers} the high resolution images. Given a collection of $n$ real samples $\{\bx_i\}_{i=1}^n$, we minimize the following reconstruction loss:
	\begin{equation}
		L_{AE}(\theta_F, \theta_H) = \frac{1}{n} \sum\limits_{i=1}^n {\left\| F\left(H(\bx_i)\right) - \bx_i \right\|_1},
	\end{equation}
	where $\theta_H$ and $\theta_F$ denote the parameters of the encoder $H$ and the decoder $F$, respectively.
	Here, we use L1 loss instead of the L2 loss to better capture the high-level and salient features~\cite{yin2008extracting}. This feature becomes the embedding which is able to uncover the primary characteristics of real data, e.g. the structural information or image style.
	In this way, the learned latent embedding has the potential to produce meaningful images.
	
	There are two advantages of extracting the embedding from data using an autoencoder.
	{Firstly}, the autoencoder extracts the high-level features which effectively preserve the primary data characteristics, \textit{e.g.}, structural information~\cite{van2009learning, salakhutdinov2007learning}, to reconstruct the original images.
	It is helpful to train a generator on the extracted features to produce meaningful samples (see results in Section~\ref{sec:exp}).
	In practice, we set the embedding to $32 \times 32$ with 64 channels to represent the $512 \times 512$ RGB images. In this sense, the dimensionality has been largely reduced by over $10$ fold
	in terms of the number of pixels.
	{Secondly}, the generator {only needs to} learn a mapping from the input noise $z \in \mathbb{R}^{100}$ to the extracted low-dimensional embedding rather than the high-dimensional images, which greatly {facilitates}
	the training of deep generative models.
	
	\begin{algorithm}[t]
		\small
		\caption{Training algorithm for AEGAN}
		\label{alg:training}
		\begin{algorithmic}[0]
			\REQUIRE  Real data $\{\bx_i\}_{i=1}^n$; prior distribution $p(\bz)$, $\bz {\small \in} \mmR^{100}$.
			\STATE Step 1: \textbf{Train autoencoder to learn low-dimensional embeddings}\\
			Update the encoder $H$ and decoder $F$ by minimizing the reconstruction loss:\\
			\qquad \qquad $L_{AE}(\theta_F, \theta_H) = \frac{1}{n} \sum\limits_{i=1}^n {\left\| F\left(H(\bx_i)\right) - \bx_i \right\|_1}$
			\STATE Step 2: \textbf{Train GANs to produce high resolution images}\\
			\STATE \qquad \textbf{for} number of training iterations \textbf{do}
			\STATE \qquad \qquad $\bullet$ Train $G_E$ and $D_E$ by optimizing the objective:\\
			\qquad \qquad \qquad \qquad ~ $\min \limits_{\theta_{G_E}} \max \limits_{\theta_{D_E}} L_{E}(\theta_{G_E}, \theta_{D_E})$ \\
			\STATE \qquad \qquad $\bullet$ Train $\phi$ and $D_R$ by optimizing the objective:\\
			\qquad \qquad \qquad \qquad  $\min \limits_{\theta_{\phi}} \max \limits_{\theta_{D_R}} L_{\phi}(\theta_{\phi}, \theta_{D_R}, \theta_{G_E}, \theta_{F})$
			\STATE \qquad \textbf{end for} 
			\STATE Step 3: \textbf{End-to-end fine-tuning} \quad \//\// optional
			\STATE Update $\mathcal{P}{\small =}\{\theta_{G_E} {\small ,} \theta_F, \theta_{\phi}, \theta_H\}, \mathcal{D}{\small =}\{\theta_{D_E}, \theta_{D_R}\}$ by optimizing:\\
			$\min \limits_{\mathcal{P}} \max \limits_{\mathcal{Q}} L_{AE}(\theta_F) {\small +} L_{E}(\theta_{G_E}, \theta_{D_E}) {\small +} L_{\phi}(\theta_{\phi}, \theta_{D_R}, \theta_{G_E}, \theta_{F})$
		\end{algorithmic}
	\end{algorithm}
	
	\subsection{Adversarial Embedding Generator}
	With the extracted embedding which contains image structural information, we {seek to} exploit it to improve the training of GANs.
	We define a generative model to match the meaningful embedding extracted from real data.
	The objective function of the generative model is as follows:
	\begin{equation}
		\begin{aligned}
			L_{E}(\theta_{G_E}, \theta_{D_E}) &= \frac{1}{n} \sum\limits_{i=1}^n { \log D_E(H(\bx_i)) } \\& + \frac{1}{n} \sum\limits_{j=1}^n { \log (1 - D_E(G_E(\bz_j))) },
		\end{aligned}
		\label{eq:embeddingGen}
	\end{equation}
	where $H(\by)$ denotes the extracted embedding from real images and $\bz$ denotes the input noise sampled from a prior distribution.
	We optimize $G_E$ and $D_E$ in an alternating manner by solving the minimax problem:
	\begin{equation}
		\min \limits_{\theta_{G_E}} \max \limits_{\theta_{D_E}}~ L_{E}(\theta_{G_E}, \theta_{D_E}).
	\end{equation}
	
	During training, we fix the parameters of autoencoder and train the GAN model to match {the distribution of embeddings} in the latent space.
	Note that we share the decoder during both extracting the high-level feature in the autoencoder and recovering the synthetic high resolution images from the generated fake embeddings.
	Ideally, with a well-trained model, we can map a noise vector to obtain a latent embedding with photo-structure information, which can be then decoded into a meaningful high resolution image.
	
	\subsection{Adversarial Denoiser Network}
	We observe that the decoded images often encounter visual noisy artifacts after going through the pipeline of the GAN and the decoder {(see Figs.~\ref{fig:network} and~\ref{fig:embedding_size_2})}. An inevitable problem is how to remove these artifacts and provide photo-realism in the generated images.
	We thus develop a denoiser network to address this issue.
	
	Actually, refining images using adversarial training has been developed in~\cite{shrivastava2016learning}, which uses {a pixel-level method}, {\it i.e.}, a chain of convolutional layers without striding or pooling to refine the blurring regions.
	However, when producing high resolution images, the synthetic images often contain severe visual artifacts which are hard to be refined using such a pixel-level method without scaling.
	{To address this issue}, we develop a new denoiser model $\phi$ {that} contains multiple layers of convolution and deconvolution operators. The structure is shown in Fig.~\ref{fig:network} (bottom).
	
	The denoiser network $\phi$ follows the encoder-decoder design.
	It is worth mentioning that the stride operation of convolution can effectively extract the primary features of {the} data and discard pixel-level noises~\cite{mao2016image}.
	To improve the fidelity of the synthetic images, we introduce an adversarial loss to provide 
	{photo-realistic details}
	in the generated images.
	We implement the $D_R$ as a convolutional network which outputs the probability of samples being positive.
	{In addition to} denoising artifacts from the generated images, we also have to preserve the content of generated images. To this end, a pixel-wise loss is taken into account.
	{As a result,} we can simultaneously preserve the image content and increase the perceptual fidelity of {the} images. The training objective becomes
	\begin{equation}
		\small
		\begin{aligned}
			L_{\phi}(\theta_{\phi}, \theta_{D_R}, \theta_{G_E}, \theta_{F}) &{\small =} \frac{1}{n} \sum\limits_{i=1}^n { \log D_R(H(\bx_i)) } \\
			&{\small +} \frac{1}{n} \sum\limits_{j=1}^n { \log (1 {\small -} D_R(\phi(\hat\bx_j))) } {\small +} \lambda \left\| \phi(\hat\bx_j) {\small -} \hat\bx_j \right\|_1,
			\label{eq:denoiserG}
		\end{aligned}
	\end{equation}
	where $\hat \bx = F(G_E(\bz))$ indicates the synthetic images before feeding them into the denoiser network.
	During training, we fix $G_E$ and $\theta_F$ and only update the adversarial denoiser network.
	{Similar to the embedding generative model}, we also train the denoiser model by solving the minimax problem:
	\begin{equation}
		\min \limits_{\theta_{\phi}} \max \limits_{\theta_{D_R}}~ L_{\phi}(\theta_{\phi}, \theta_{D_R}, \theta_{G_E}, \theta_{F}).
	\end{equation}
	
	{With the denoiser network, the noisy artifacts can be effectively removed from the generated images. As a result, the proposed method is able to produce promising images with high perceptual fidelity.
	}
	
	\subsection{Training and Inference Method}
	The proposed method consists of three components, namely the autoencoder, the embedding generative model and the denoiser network. To effectively train the whole model, we {adopt a step-wise strategy to train all the components} and then conduct end-to-end fine-tuning to obtain a new unified model. The training procedure is shown in Algorithm~\ref{alg:training}.
	
	We divide the training of the proposed AEGAN into three steps.
	In the first step, we train the autoencoder by minimizing the reconstruction loss to extract the low-dimensional embedding. The {learned} embedding is able to capture {the} image structural information and effectively recover the high resolution images.
	In the second step, by fixing the autoencoder model, we train an embedding generative model and a denoiser network in an {alternating} manner. In the proposed model, the autoencoder acts as a bridge {that} connects all the three components.
	
	However, {when} one of the components fails, there is no way to correct it. To ensure the final performance and obtain a new unified model, in the last step, we conduct end-to-end training to fine-tune the whole model by optimizing the following joint loss:
	\begin{equation}
		\small
		\min \limits_{\mathcal{P}} \max \limits_{\mathcal{Q}}~ L_{AE}(\theta_F, \theta_H) {\small +} L_{E}(\theta_{G_E}, \theta_{D_E}) {\small +} L_{\phi}(\theta_{\phi}, \theta_{D_R}, \theta_{G_E}, \theta_{F}),
		\label{eq:joint-finetuning}
	\end{equation}
	where $\mathcal{P} =\{\theta_{G_E}, \theta_F, \theta_{\phi}, \theta_H\}$ 
	and $\mathcal{Q}=\{\theta_{D_E}, \theta_{D_R}\}$.
	In each iteration of the final step, we first update the discriminators $D_E$ and $D_R$ using the gradients propagated from $L_E$ and $L_{\phi}$, respectively.
	{We then}
	backpropagate the signals of each loss through the model and the gradient for each component will accumulate.
	The fine-tuning step ensures the coherence of the whole model and is able to achieve slightly better performance. 
	In practice, we observe that failure of {the} components rarely occurs. {Thus,} the fine-tuning step {is} optional.
	Compared to directly optimizing the objective in Eqn.~(\ref{eq:joint-finetuning}), each separately trained model can be viewed as a good initialization {and can be used} to accelerate the training to obtain a new unified model.
	
	For inference, 
	we sample a vector $\bz \in \mmR^{100}$ from a prior distribution to produce a latent embedding. 
	{We then} feed the latent embedding into the decoder $F$ to produce a corresponding high resolution image.
	{Finally, we} take the generated images into the denoiser model. Through the pipeline of matching {the} random noise to a high resolution image, AEGAN is able to generate $512 \times 512$ RGB images with promising quality in terms of both quantitative and qualitative results. 
	
	\begin{table*}[htbp]
		\centering
		\caption{Comparisons of various generative models in terms of FID, Inception Score and multi-scale structural similarity (MS-SSIM) on Oxford-102, CUB, CelebA-HQ, LSUN and ImageNet datasets. We use $10,000$ samples for testing. Higher inception score and lower FID indicate better image quality. Lower MS-SSIM scores indicate higher diversity.}
		\begin{tabular}{c|c|c|c|c|c|c|c|c}
			\hline
			\multirow{3}[0]{*}{Methods} & \multicolumn{8}{c}{FID} \\
			\cline{2-9}
			& \multirow{2}[0]{*}{Oxford-102} & \multirow{2}[0]{*}{CUB} & \multirow{2}[0]{*}{CelebA-HQ} & \multicolumn{3}{c|}{LSUN} & \multicolumn{2}{c}{ImageNet} \\
			\cline{5-9}          &       &       &       & Bedroom & Classroom & Conference & Volcano & Promontory \\
			\hline
			DCGAN~\cite{radford2015unsupervised} & 76.19 & 184.09 & 37.87 & 145.32 & 401.64 & 292.05 & 128.01 & 246.60 \\
			\hline
			WGAN-GP~\cite{gulrajani2017improved} & 139.90 & 171.33 & 55.93 & 132.05 & 161.10 & 130.97 & 154.29 & 169.72 \\
			\hline
			AGE~\cite{ulyanov2017adversarial}   & 311.33 & 261.15 & 94.42 & 175.49 & 245.45 & 230.90 & 280.91 & 292.74 \\
			\hline
			StackGAN~\cite{zhang2016stackgan} & 168.79 & 155.98 & 36.63 & 161.62 & 243.63 & 285.33 & 161.95 & 158.43 \\
			\hline
			StackGAN-v2~\cite{Han17stackgan2} & 88.18 & 90.64 & 33.17 & 71.09 & 446.44 & 78.27 & 78.72 & 89.35 \\
			\hline
			Progressive GAN~\cite{karras2017progressive} & 65.31 & 87.57 & \textbf{18.64} & 36.02 & \textbf{108.40} & 42.52 &    30.01   & 41.39 \\
			\hline
			AEGAN (ours) &   \textbf{64.51}    &   \textbf{83.74}    &    18.83   &   \textbf{34.97}    &  108.52     &    \textbf{41.16}   &    \textbf{28.47}   & \textbf{33.24} \\
			\hline
			\hline
			& \multicolumn{8}{c}{Inception Score} \\
			\hline
			DCGAN~\cite{radford2015unsupervised} & 3.27$\pm$0.08 & 4.43$\pm$0.09 & \textbackslash{} & 3.25$\pm$0.06 & 1.02$\pm$0.01 & 1.92$\pm$0.02 & 2.14$\pm$0.05 & 1.83$\pm$0.02 \\
			\hline
			WGAN-GP~\cite{gulrajani2017improved} & 3.52$\pm$0.07 & 4.51$\pm$0.15 & \textbackslash{} & 3.63$\pm$0.03 & 2.86$\pm$0.06 & 3.37$\pm$0.06 & 2.66$\pm$0.04 & 1.71$\pm$0.03 \\
			\hline
			AGE~\cite{ulyanov2017adversarial}   & 2.33$\pm$0.06 & 2.70$\pm$0.03 & \textbackslash{} & 3.41$\pm$0.05 & 2.58$\pm$0.03 & 2.80$\pm$0.02 & 1.66$\pm$0.02 & 1.78$\pm$0.02 \\
			\hline
			StackGAN~\cite{zhang2016stackgan} & 3.38$\pm$0.07 & 4.13$\pm$0.08 & \textbackslash{} & 2.99$\pm$0.06 & 1.84$\pm$0.03 & 1.83$\pm$0.01 & 2.64$\pm$0.07 & 2.18$\pm$0.04 \\
			\hline
			StackGAN-v2~\cite{Han17stackgan2} & 3.92$\pm$0.08 & 4.45$\pm$0.08 & \textbackslash{} & 3.02$\pm$0.09 & 1.79$\pm$0.02 & 3.57$\pm$0.07 & 1.97$\pm$0.05 & 2.12$\pm$0.03 \\
			\hline
			Progressive GAN~\cite{karras2017progressive} & 3.83$\pm$0.10 & 4.80$\pm$0.18 & \textbackslash{} & 3.50$\pm$0.03 & \textbf{3.56$\pm$0.11} & 3.56$\pm$0.08 &   2.73$\pm$0.03    & 2.14$\pm$0.04 \\
			\hline
			AEGAN (ours) & \textbf{3.98$\pm$0.07} & \textbf{4.85$\pm$0.09} & \textbackslash{} &   \textbf{3.57$\pm$0.07}    &   {3.41$\pm$0.03}    &   \textbf{3.61$\pm$0.05}    & \textbf{2.82$\pm$0.02}    & \textbf{2.21$\pm$0.03} \\
			\hline
			\hline
			& \multicolumn{8}{c}{MS-SSIM} \\
			\hline
			DCGAN~\cite{radford2015unsupervised} & 0.3196 & 0.4295 & 0.3414 & 0.2578 & 0.9608 & 0.5725 & 0.3391 & 0.3541 \\
			\hline
			WGAN-GP~\cite{gulrajani2017improved} & 0.2710 & 0.3404 & 0.2868 & 0.1903 & 0.1136 & 0.1554 & 0.3031 & 0.2804 \\
			\hline
			AGE~\cite{ulyanov2017adversarial}   & 0.2829 & 0.3568 & 0.2878 & 0.1993 & 0.1908 & 0.2031 & 0.3379 & 0.3647 \\
			\hline
			StackGAN~\cite{zhang2016stackgan} & 0.2674 & 0.3286 & 0.3504 & 0.3368 & 0.5994 & 0.4786 & 0.3199 & 0.3300 \\
			\hline
			StackGAN-v2~\cite{Han17stackgan2} & 0.2379 & 0.3193 & 0.2948 & 0.3021 & 0.7073 & 0.2711 & 0.4328 & 0.4023 \\
			\hline
			Progressive GAN~\cite{karras2017progressive} & \textbf{0.2184} & \textbf{0.2927} & 0.2934 & 0.1861 & \textbf{0.1483} & 0.1477 &   0.3087    & 0.2912 \\
			\hline
			AEGAN (ours) & 0.2310 & 0.3125 &   \textbf{0.2927}    &   \textbf{0.1794}    &   0.1532    &   \textbf{0.1435}    &    \textbf{0.3011}   & \textbf{0.2785} \\
			\hline
		\end{tabular}
		\label{tab:fid}
	\end{table*}
	
	\begin{figure*} [t]
		\centering
		\includegraphics[width=2\columnwidth]{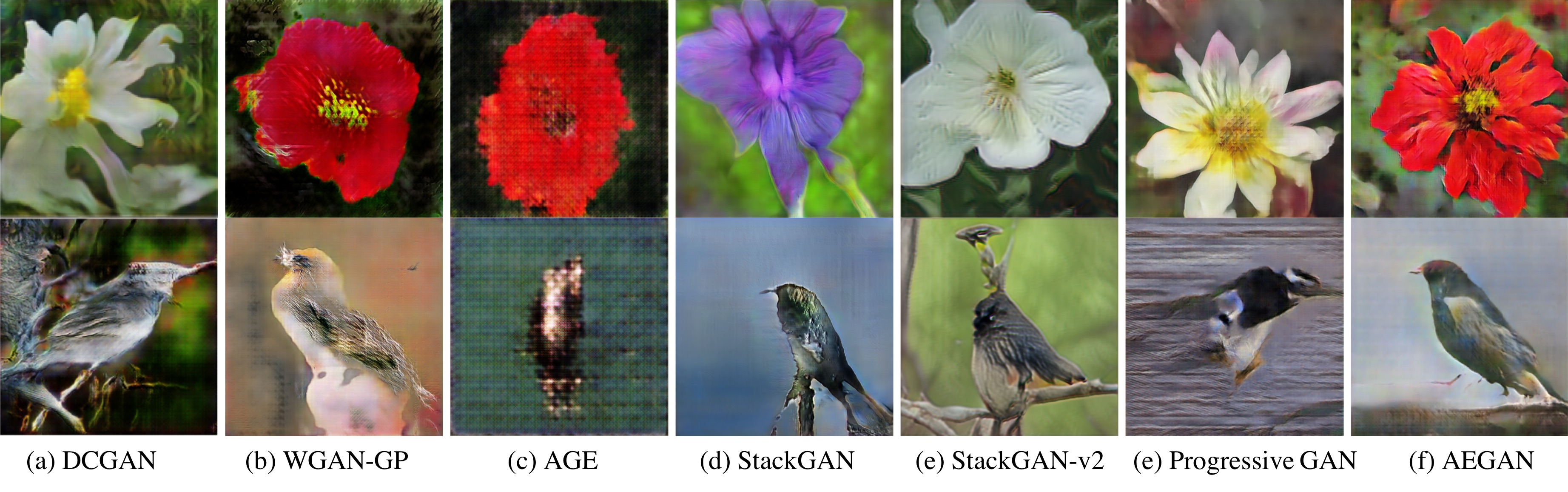}
		\caption{Visual comparison of $512 \times 512$ {images} generated {using} different methods on {the} Oxford-102 (top) and CUB (bottom) datasets.}
		\label{fig:flower-bird}
	\end{figure*}
	
	\subsection{{Model Architecture}}
	{
		We build the embedding generator $G_E$ and the decoder $F$ with a stack of up-sampling blocks (\emph{i.e.}, a strided deconvolutional layer followed by a residual blocks~\cite{he2016deep}). The number of blocks is determined by the upscaling factor of {the} feature maps.
		In the experiment, we add 3 and 4 up-sampling blocks in $G_E$ and {$F$}, respectively. 
		For the encoder {$H$}, we build the model with a stack of down-sampling blocks to encode RGB images to the low-dimensional embeddings. Each down-sampling block contains a strided deconvolutional layer followed by a residual blocks~\cite{he2016deep}. Similar to the design of the decoder $F$, we also add 4 down-sampling blocks in $H$.
		For the discriminators $D_R$ and $D_E$, we build the model with a stack of strided convolution layers to down-sample the input images or embeddings.
		We {add} a Batch Normalization~\cite{ioffe2015batch} and a ReLU~\cite{nair2010rectified} layer behind each convolutional layer.
		The denoiser network $\phi$ follows the encoder-decoder design. The input generated images are fed into several down-sampling modules (\emph{i.e.}, a strided convolutional layer followed by a Batch Normalization and a LeakyReLU~\cite{maas2013rectifier} layer) until it has a size of $512$. A series of up-sampling modules (\emph{i.e.}, strided deconvolutional layer followed by a Batch Normalization and a ReLU layer) are {then} used to generate $512\times512$ RGB high-resolution images. 
		The strides of all the strided convolutional {and} deconvolution layer are set to 2 and 1/2, {respectively}.
	}
	
	\subsection{Implementation Details}
	We follow the experimental settings {used} in~\cite{radford2015unsupervised} to train the proposed AEGAN in PyTorch.
	There is no preprocessing or data augmentation {of the} training samples {other than} resizing all {the} training images to $512 \times 512$ RGB images.
	In all the experiments, we set the embedding to $32 \times 32$ with 64 channels to represent the $512 \times 512$ RGB images.
	The weights are initialized from a normal distribution with zero-mean and standard deviation of 0.02.
	The hyperparameter $\lambda$ in Eqn. (\ref{eq:denoiserG}) is set to $\lambda=100$. We use ReLU activation in the upsamplers, such as the embedding generator in GANs and the decoder in the autoencoder, while using LeakyReLU in the downsamplers including discriminators in GANs and the encoder in the autoencoder. The slope of the leak in LeakyReLU is set to $0.2$.
	In the training, we use Adam~\cite{kingma2014adam} with $\beta_1=0.9$ to update the model parameters.
	In the first {training} step, we use a fixed learning rate of $10^{-5}$ to train the autoencoder.
	For the second step, we follow the same settings {used in} DCGAN~\cite{radford2015unsupervised} and set the learning rate to $0.0002$. At the final step, we adjust the learning rate to $10^{-7}$ for end-to-end fine-tuning. {During training, we set the minibatch size to $16$. For the three steps in Algorithm~\ref{alg:training}, we train the corresponding models with $100$ epochs, $200$ epochs and $100$ epochs, respectively.}
	
	\section{Experiments}\label{sec:exp}
	In the experiments, we focus on generating high resolution 
	images of $512 \times 512$.
	To evaluate the performance of the proposed method, several generative models are adopted for comparison, including DCGAN~\cite{radford2015unsupervised},
	WGAN-GP~\cite{gulrajani2017improved},
	AGE~\cite{ulyanov2017adversarial}, 
	StackGAN~\cite{zhang2016stackgan},
	{StackGAN-v2~\cite{Han17stackgan2}}
	and Progressive GAN~\cite{karras2017progressive}.
	
	{For the comparison methods, we use the official source codes and the original settings {specified in} their papers. Specifically, we train DCGAN, WGAN-GP, AGE and StackGAN-v2 for 200, 3906, 150 and 600 epochs, respectively, following the default settings in the original papers.
		We train StackGAN for 220 epochs, \emph{i.e.}, 120 for Stage-I and 100 for Stage-II. For Progressive GAN, we start with $4 \times 4$ resolution and train the model at each resolution using 1200k images, \emph{i.e.}, 600k for fade-in and 600k for stabilizing.}
	
	For convenience, we organize the experiments as follows. Firstly, we introduce some details about the benchmark datasets and the evaluation metrics in Section~\ref{exp:dataset}. Then, we compare the performance of our method with several baselines on five benchmark datasets and show {both quantitative and {qualitative} comparisons in Section~\ref{exp:quantitative} and~\ref{exp:visual}.} 
	
	\begin{figure*}[t]
		\centering
		\includegraphics[trim = 1mm 0mm 0mm 0mm,
		clip, width=2\columnwidth]{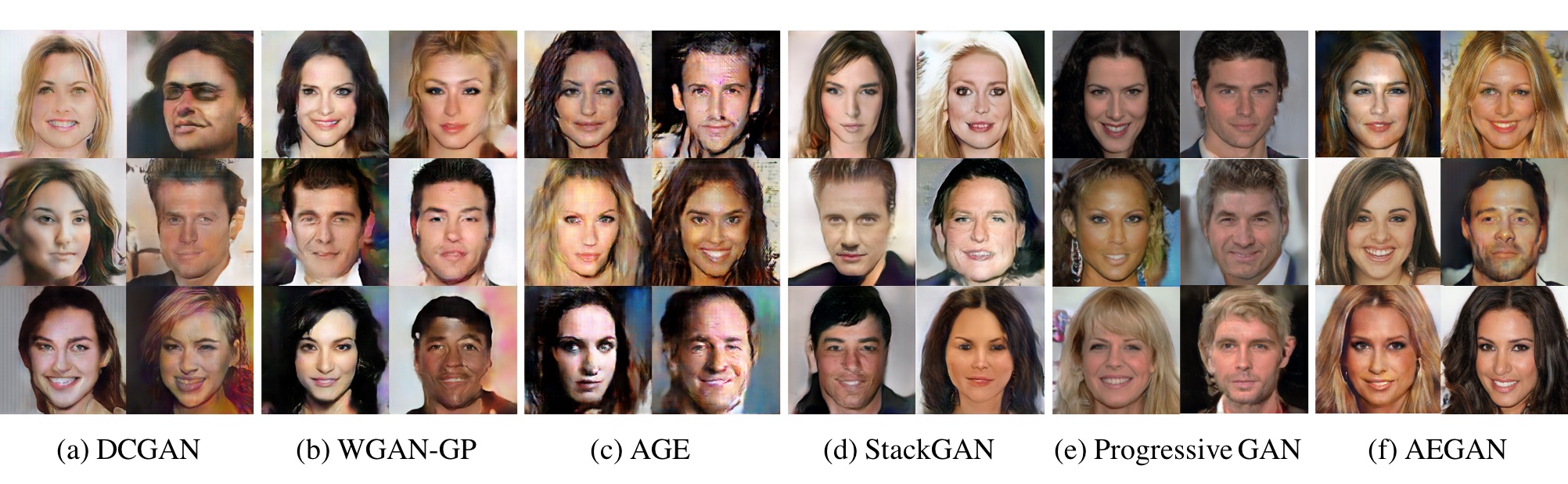}
		\caption{Visual comparison of $512 \times 512$ face examples produced by different generative models on CelebA-HQ dataset. {The samples generated by each method are chosen randomly.}}
		\label{fig:face1}
	\end{figure*}
	
	\begin{figure*} [t]
		\centering
		\includegraphics[width=2\columnwidth]{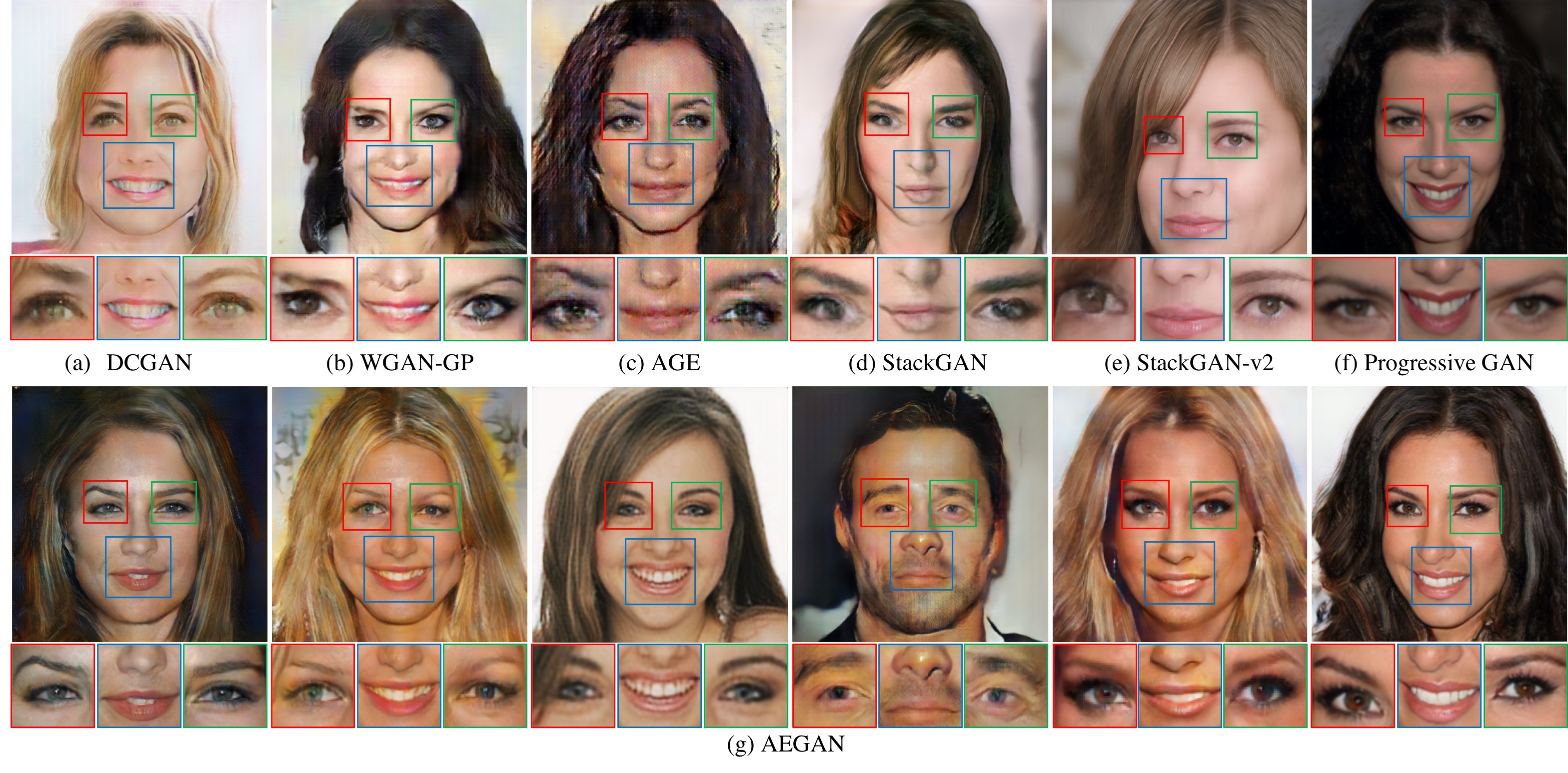}
		\caption{Visual comparison of high resolution {images of} face  generated {by various} methods on {the} CelebA-HQ dataset. We compare the detailed structure and texture including {the} eyes, nose, and mouth. The top row shows the results of different baselines; the bottom row shows the results of the proposed AEGAN.}
		\label{fig:face_hq}
	\end{figure*}
	
	\subsection{Datasets and Evaluation Metrics}\label{exp:dataset}
	We evaluate the proposed method {using} a wide variety of benchmark datasets, including Oxford-102 Flowers~\cite{nilsback2008automated}, Caltech-UCSD Birds (CUB)~\cite{wah2011caltech}, High-Quality Large-scale CelebFaces Attributes (CelebA-HQ)~\cite{liu2015faceattributes,karras2017progressive}, Large-scale Scene Understanding (LSUN)~\cite{yu15lsun} and ImageNet~\cite{russakovsky2015imagenet}.
	Oxford-102 Flowers contains $8189$ images of flowers from 102 fine-grained classes and CUB contains $200$ bird species with $11,788$ high resolution images in total. 
	{The images in Oxford-102 Flowers have resolutions of over $500 \times 500$ while CUB {contains} images with resolution ranging from $300$ to $500$.}
	{{The} CelebA-HQ dataset,which is generated from the original CelebA dataset~\cite{liu2015faceattributes}, contains 30K celebrity face images at $1024 \times1024$ {resolution}.}  
	{In our experiments, we resize the training samples to $512 \times 512$ resolution.}
	LSUN contains {approximately} one million images {at} the resolutions ranging from $300$ to $500$. 
	{ImageNet contains 1,000 classes {and} 1.28 million images with {the resolutions}  ranging from $200$ to $700$ (the average image resolution is $469\times387$).  Due to the setting of unconditional image generation, we train all the methods on each category separately.}
	For each dataset, we resize all the images to $512 \times 512$ resolution {during} training.
	
	For quantitative evaluation, we use {\emph{{Frechet} Inception Distance (FID)}~\cite{heusel2017gans}}, \emph{Inception Score}~\cite{gulrajani2017improved} and \emph{MS-SSIM}~\cite{odena2016conditional,wang2003multiscale} to evaluate the generated samples.
	{FID is a widely used metric for implicit generative models, as it correlates well with the visual quality of generated samples.}
	Inception Score {can be used} to measure both image quality and {image} diversity over a large number of samples. In general, larger {Inception Score} value {indicates} the better performance. MS-SSIM measures the diversity of {the} generated samples and the {resulting} values range from 0.0 to 1.0. Higher MS-SSIM values correspond to perceptually more similar images. 
	{In the experiments, we use $10,000$ images to calculate these scores. For Inception Score, 10 splits are used to compute the standard deviations.}
	
	\begin{figure*} [t]
		\centering
		\includegraphics[width=2\columnwidth]{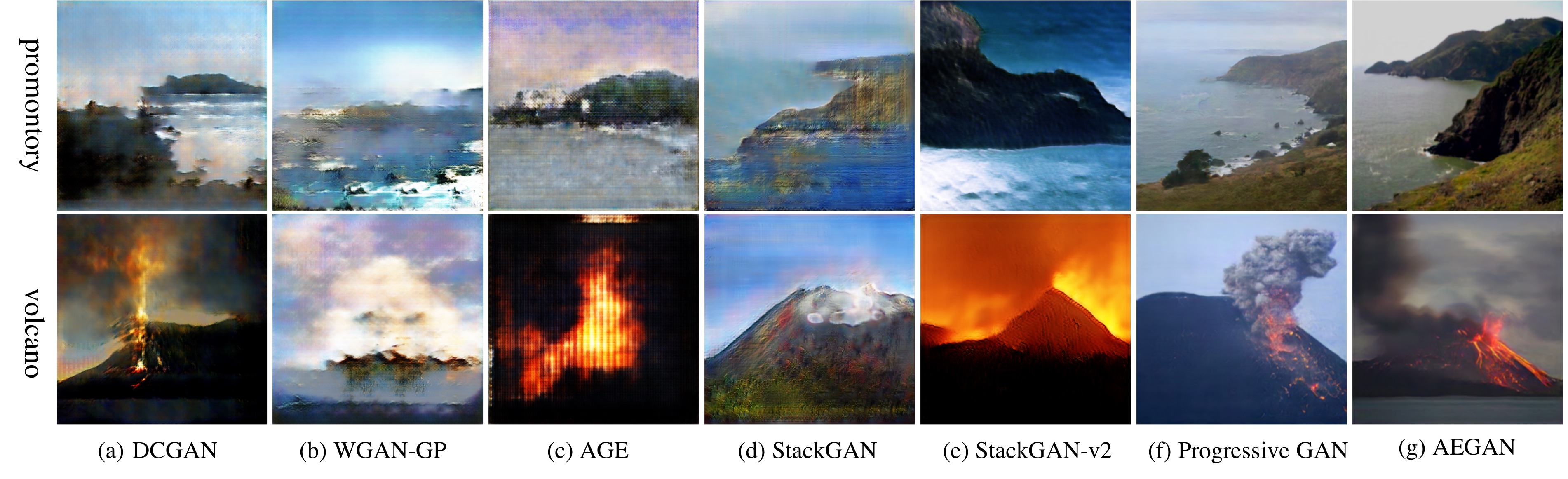}
		\caption{Visual comparison of high resolution examples generated {using} different methods on the ImageNet dataset. For convenience, we choose two categories, \emph{i.e.}, Promontory and Volcano, to show the results.}
		\label{fig:imagenet}
	\end{figure*}
	
	\begin{figure*}[t]
		\centering
		\includegraphics[trim = 1mm 0mm 0mm 0mm,
		clip, width=2\columnwidth]{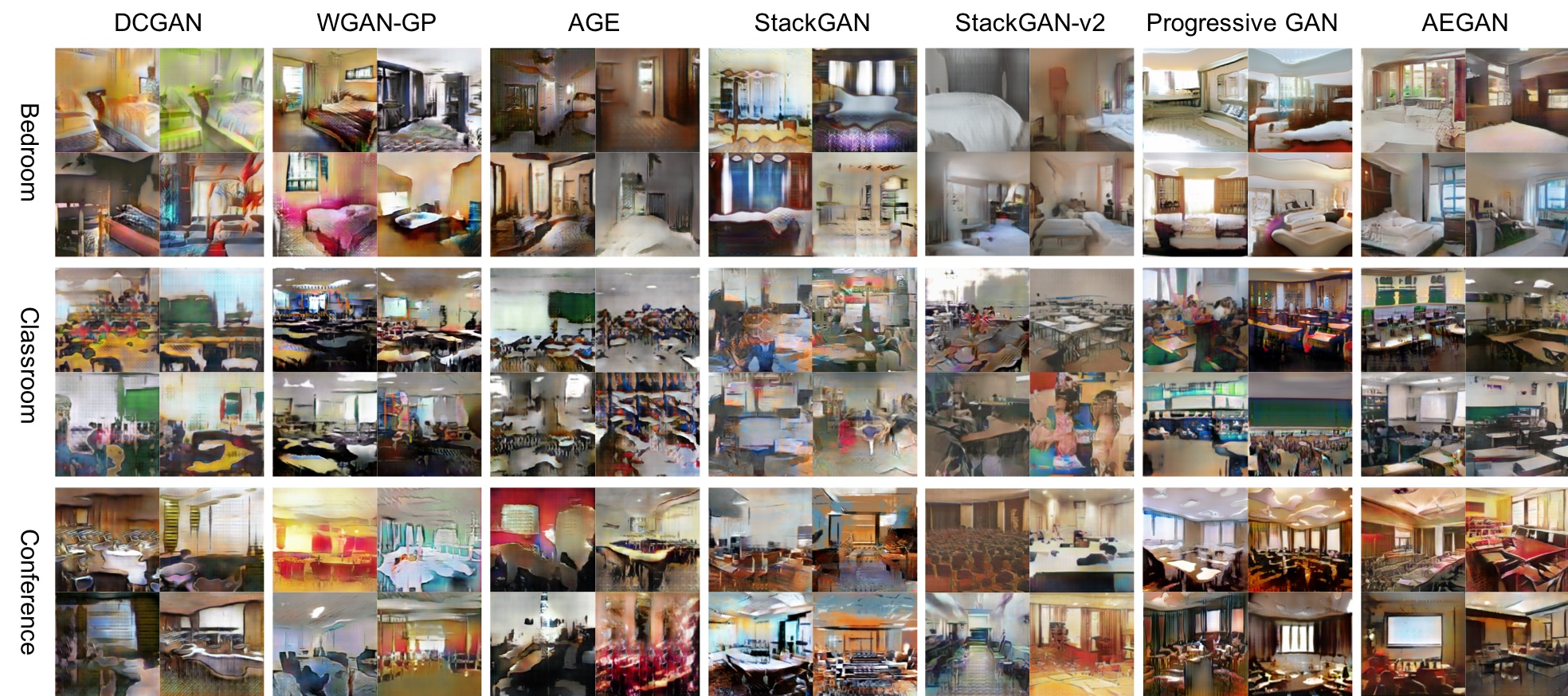}
		\caption{Visual comparison of $512 \times 512$ {images produced by}  different generative models on the LSUN datasets. We randomly choose  samples generated {using each} method for careful observation.}
		\label{fig:lsun}
	\end{figure*}
	
	\subsection{Quantitative Comparisons}\label{exp:quantitative}
	
	\begin{table*}[t]
		\centering
		\caption{Demonstration of the effect of denoiser network on Oxford-102 dataset. We compare AEGAN with 6 baseline methods with and without the denoiser module in terms of FID, Inception Score and MS-SSIM.}
		\begin{tabular}{c|c|c|c|c|c|c}
			\hline
			\multicolumn{1}{c|}{\multirow{2}[0]{*}{Methods}} 
			& \multicolumn{2}{c|}{FID} & \multicolumn{2}{c|}{Inception Score} & \multicolumn{2}{c}{MS-SSIM}  \\
			\cline{2-7}
			& \multicolumn{1}{c|}{original} & \multicolumn{1}{c|}{with denoiser} & \multicolumn{1}{c|}{original} & \multicolumn{1}{c|}{with denoiser} & \multicolumn{1}{c|}{original} & \multicolumn{1}{c}{with denoiser} \\
			\cline{1-7}
			DCGAN~\cite{radford2015unsupervised} &   76.19    &   73.02  &   $3.27 \pm 0.08$    &   $3.45 \pm 0.05$    &   0.3196   &   0.3445     \\
			WGAN-GP~\cite{gulrajani2017improved} &   139.90    &    135.73   &  $3.52 \pm 0.07$     &    $3.74 \pm 0.09$   &   0.2710    &   0.2859    \\
			AGE~\cite{ulyanov2017adversarial}   &   311.33    &   287.94   &   $2.33 \pm 0.06$    &    $3.56 \pm 0.09$   &   0.2829    &   0.3108    \\
			StackGAN~\cite{zhang2016stackgan} &   168.79    &    153.67   &   $3.38 \pm 0.07$    &   $3.86 \pm 0.10$    &   0.2674    &   0.2715    \\
			StackGAN-v2~\cite{Han17stackgan2} &    88.18   &    87.30   &  $3.92 \pm 0.08$  &  $3.94 \pm 0.10$  &  0.2379 &  0.2404 \\
			Progressive GAN~\cite{karras2017progressive} &    65.31   &   65.07    &   $3.83 \pm 0.10$    &   $3.87 \pm 0.09$   &   \textbf{0.2184}    &  0.2237     \\
			\hline
			AEGAN &      \multicolumn{2}{c|}{\textbf{64.51}}   & \multicolumn{2}{c|}{\textbf{3.98 $\pm$ 0.07}}     &      \multicolumn{2}{c}{0.2310}    \\
			\hline
		\end{tabular}
		\label{tab:denoiser}
	\end{table*}
	
	\begin{figure*}[t]
		\centering
		\includegraphics[width=2.0\columnwidth]{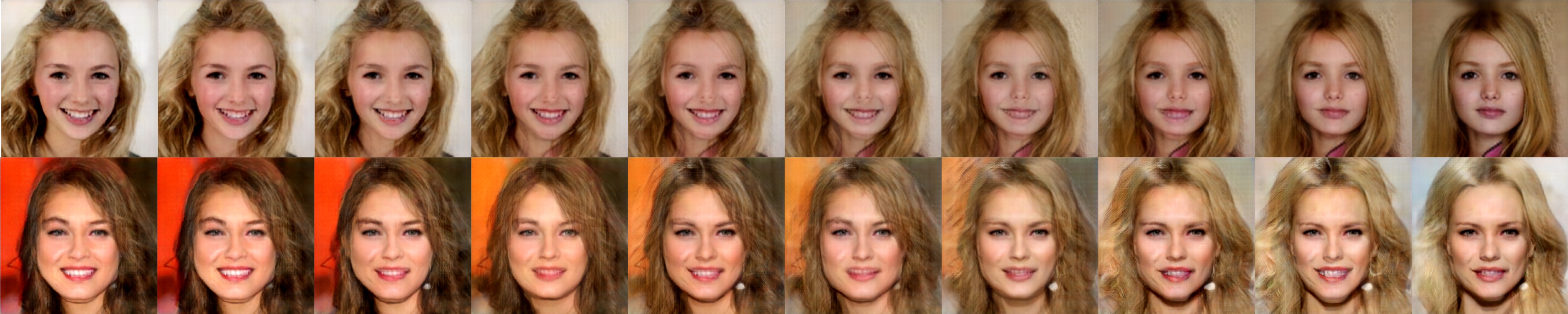}
		\caption{Results of latent space interpolation by AEGANs on the CelebA-HQ dataset. For each row, we conduct linear interpolation between two data points sampled from the prior distribution $p(\bz)$. The leftmost and rightmost columns are the generated images of AEGAN, and the columns 2 to 9 show the interpolated images.}
		\label{fig:interpolation}
	\end{figure*}

	{
		In this section, we compare the performance of different methods {on 5 benchmark datasets} in terms of FID~\cite{heusel2017gans}, Inception Score~\cite{gulrajani2017improved} and MS-SSIM~\cite{odena2016conditional,wang2003multiscale}.
		The results are shown in Table~\ref{tab:fid}.
	}
	
	{
		{On the Oxford-102 and CUB datasets,} 
		{the proposed AEGAN model obtains slightly worse results than Progressive GAN~\cite{karras2017progressive} in terms of MS-SSIM but achieves the best FID and Inception Score.}
		{On the CelebA-HQ dataset, AEGAN still obtains comparable performance with the strong baseline Progressive GAN}. 
		These results demonstrate that the proposed method is able to produce images {of} promising quality while {maintaining} a large diversity. 
	}
	
	{For the more challenging datasets LSUN and ImageNet, we train AEGAN and the considered baseline methods on each category separately. On LSUN, we compare the results {in} 3 categories, including Bedroom, Classroom and Conference. However, on ImageNet (containing 1000 categories in total), training 1000 models is infeasible and impractical. For convenience, we choose two of them, \textit{i.e.}, Promontory and Volcano, to evaluate the performance of our proposed method. From Table~\ref{tab:fid}, AEGAN outperforms the considered baseline methods {in} most categories in terms of FID, Inception score and MS-SSIM.
		These results demonstrate the superiority of the proposed method in high resolution image synthesis.
	}
	
	\subsection{Qualitative Comparisons}\label{exp:visual}
	{
		In this section, we compare the visual quality of the images generated by the proposed AEGAN {and several baseline methods} on 5 benchmark datasets, including Oxford-102, CUB, CelebA-HQ, LSUN and ImageNet.
	}
	
	{
		Figs.~\ref{fig:flower-bird} and \ref{fig:face1} {show that,} when producing high resolution images of $512 \times 512$, DCGAN tends to produce colorful but meaningless images {that} contain {many} isolated regions. 
		WGAN-GP~\cite{gulrajani2017improved} and AGE~\cite{ulyanov2017adversarial} are able to capture the rough structure and produce meaningful images. However, the generated samples {appear} very blurred and lack photo-realistic details. 
		Note that StackGAN is originally devised to generate images {by taking} the input text as a condition. 
		In this experiment, since there is no additional information acting as the condition,
		the transition of StackGAN from the first stage to the second one will be incoherent. {As a result}, the model {fails to effectively} capture the style and texture information.
		Compared to Progressive GAN~\cite{karras2017progressive}, AEGAN is able to produce promising images with {sharper} object structure and richer details than the other methods.
	}
	
	{
		On CelebA-HQ, we also compare some detailed regions of the generated images in Fig.~\ref{fig:face_hq}. Specifically, we compare the detailed structure and texture in the generated face images, such as eyes, nose, and mouth. 
		{The top row shows 
			{the images generated by the five considered baselines and their detailed regions,}
			and the bottom row shows a set of {samples generated by} AEGAN.}
		Compared to the considered baselines, 
		{AEGAN is able to produce promising images with sharper {facial} structure and finer details.}
	}
	
	{
		When generating more complex images on LSUN and ImageNet, AEGAN is able to produce meaningful images of better {perceptual} quality compared to the baseline methods. Figs.~\ref{fig:imagenet} and~\ref{fig:lsun} {show that none of} the baseline methods except Progressive GAN, is able to produce visually acceptable scene images. 
		{The images generated by Progressive GAN also lack}
		some structural and textural information. However, most of the images generated by AEGAN capture the salient features of the specific category and 
		{obtain objects with complete and sharp structure,}
		e.g. the structure of beds and the shape of desks. {This} is attributed to the extracted high-level embedding. The images generated by AEGAN are more photo-realistic than the images generated by the considered baselines in all categories.
		These results demonstrate the effectiveness of the proposed AEGAN {in} producing high resolution images.
	}
	
	\begin{figure} [t]
		\begin{center}
			\subfigure[Failure case.]{
				\includegraphics[width=0.45\columnwidth]{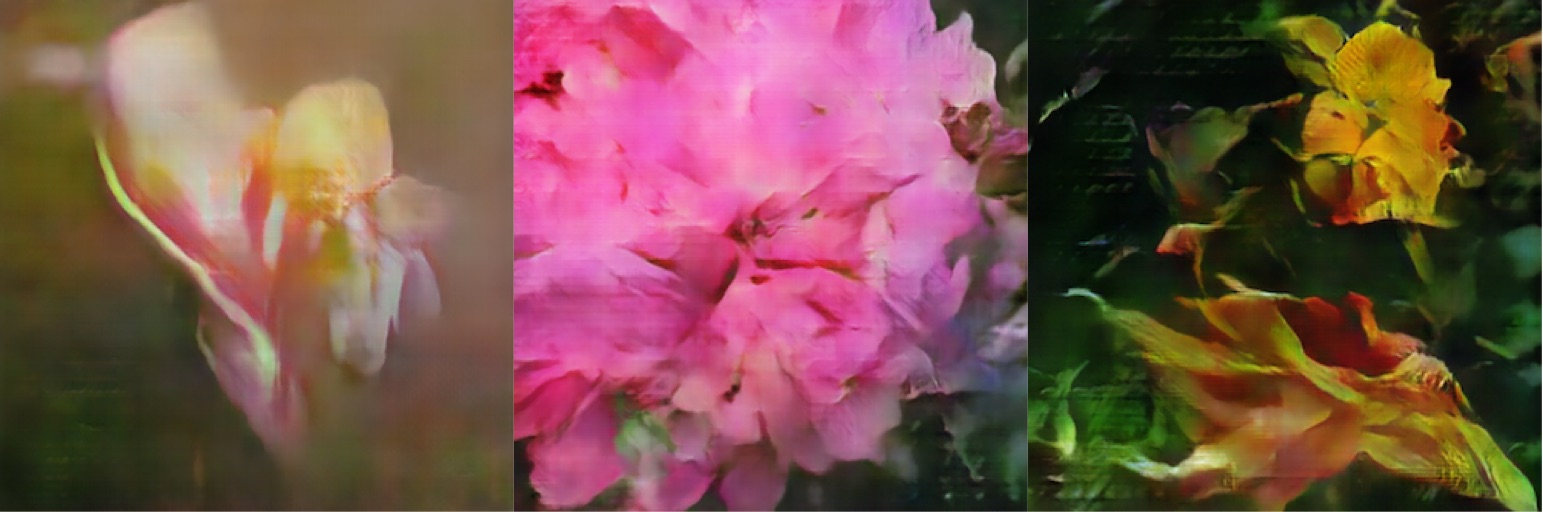}
			}
			\subfigure[AEGAN results.]{
				\includegraphics[width=0.45\columnwidth]{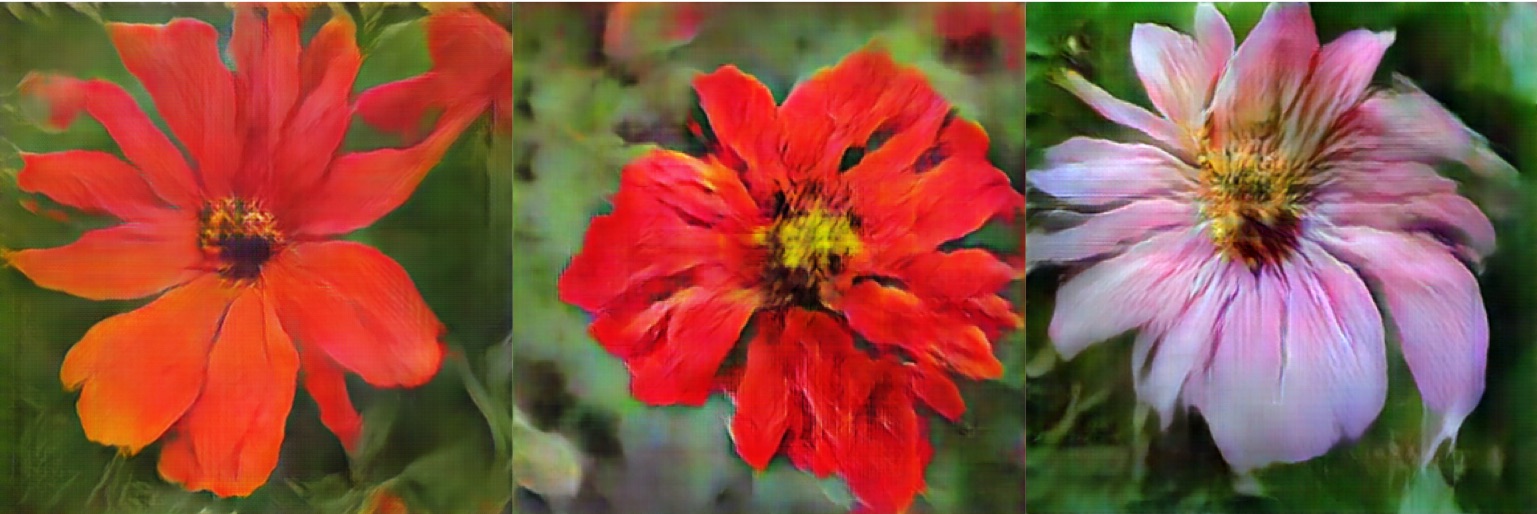}
			}
		\end{center}
		\caption{Demonstration of the possible failure case without end-to-end fine-tuning on the Oxford-102 dataset.}
		\label{fig:failure}
	\end{figure}
	
	\section{{Further Experiments}}\label{sec:discussions}
	
	{In this section, we conduct further analyses and discussions {of} our proposed method.
		{In Section~\ref{exp:ablation_finetuning}}, we perform an ablation study of the end-to-end fine-tuning step.
		{In section~\ref{effect_denoiser}, we discuss the effect of our proposed denoiser networks.}
		{In Section~\ref{exp:embeddingDim}}, we investigate the effect of embedding dimension on the performance of AEGAN.
		{In Section~\ref{exp:interpolation}}, we conduct an experiment of latent space interpolation.
	}
	{Finally, we compare the training time of our AEGAN to baselines in Section~\ref{exp:training_time}.}
	
	\subsection{{Ablation} Study on End-to-end Fine-tuning}\label{exp:ablation_finetuning}
	In this section, we {investigate} the effect of the final fine-tuning step (the third step in Algorithm~\ref{alg:training}) in generating high resolution images by ablation study. The experimental results are shown in Table~\ref{tab:abalation_finetuning}.
	From Table~\ref{tab:abalation_finetuning}, the model with fine-tuning achieves slightly better performance than the model without fine-tuning.
	
	{
		We note that there is a risk of model failure in the first two steps. Fortunately, the end-to-end fine-tuning is able to effectively address this issue, {thereby producing} 
		a new unified model with better performance. To verify this, we show some examples of a failure case in Fig.~\ref{fig:failure}. However, the failure rarely occurs in practice. The end-to-end fine-tuning helps to guarantee the performance of our method.
	}
	
	\subsection{Effect of Denoiser}\label{effect_denoiser}
	{
		In this section, we investigate the effect of the denoiser module on the performance of high resolution images synthesis. To this end, we add the denoiser module to all the considered baselines and compare {the results obtained using these}
		methods with or without the denoiser module. The experimental results are shown in Table~\ref{tab:abalation_finetuning}.
		From Table~\ref{tab:abalation_finetuning}, the denoiser module is able to improve the performance of different methods in terms of both FID and Inception Score, {demonstrating} the effectiveness of the proposed denoiser network. 
		Moreover, AEGAN consistently outperforms the baseline methods with or without the denoiser network. 
	}
	
	\begin{table}[tbp]
		\large
		\centering
		\caption{Performance comparison of AEGAN with or without the end-to-end fine-tuning on Oxford-102 dataset.}
		\resizebox{0.49\textwidth}{!}
		{
			\begin{tabular}{c|c|c}
				\hline
				Samples & \includegraphics[width=0.45\columnwidth]{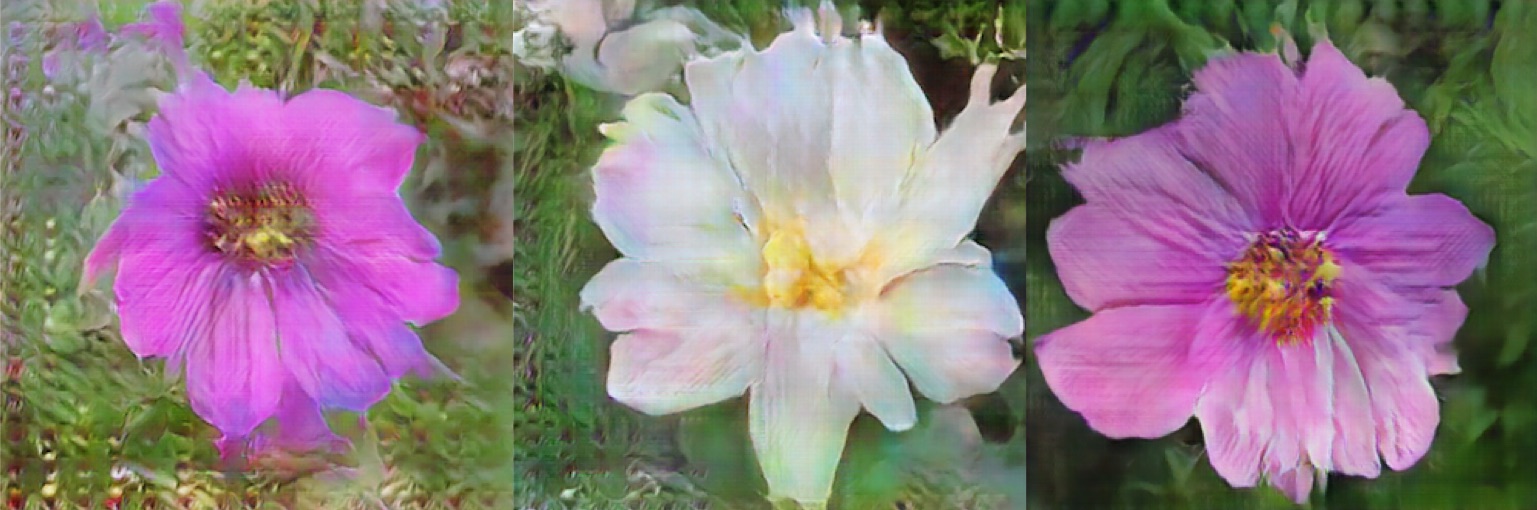} & \includegraphics[width=0.45\columnwidth]{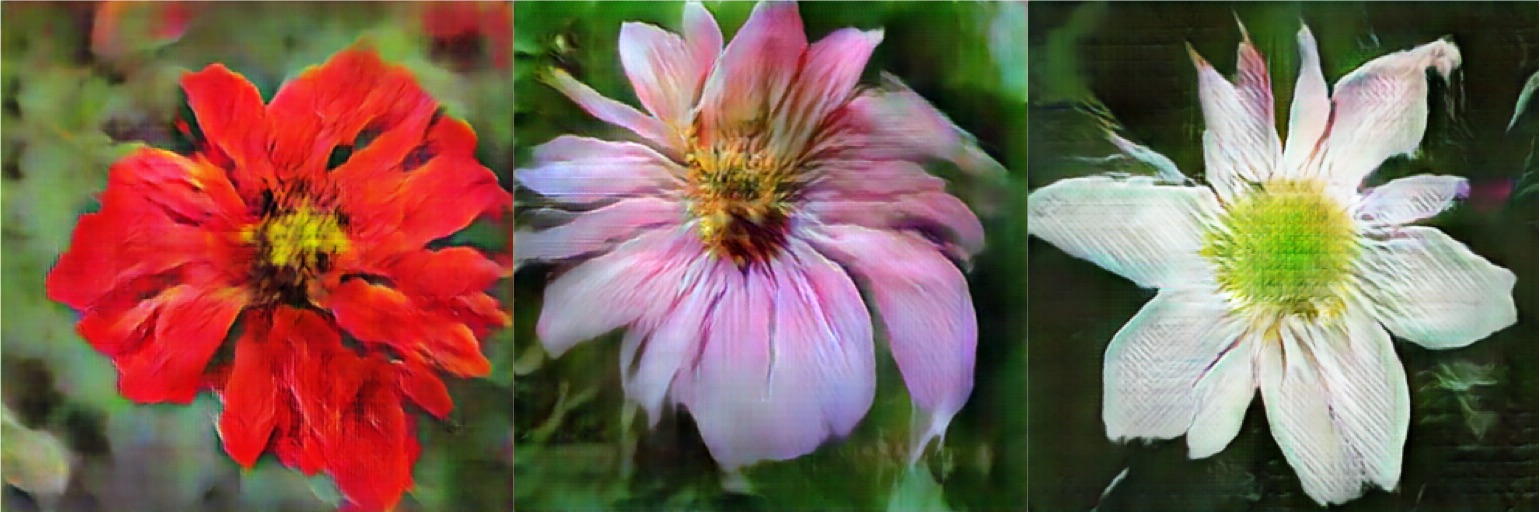} \\
				\hline
				Methods & AEGAN without fine-tuning & AEGAN with fine-tuning \\
				\hline
				FID & 64.56 &   64.51    \\
				Inception Score & 3.91 $\pm$ 0.08 &   3.98 $\pm$ 0.07    \\
				MS-SSIM & 0.2342 &  0.2310     \\
				\hline
			\end{tabular}
		}
		\label{tab:abalation_finetuning}
	\end{table}
	
	\subsection{Effect of Embedding Dimension}\label{exp:embeddingDim}
	
\begin{figure}[t]
		\centering
		\subfigure[$8 \times 8 \times 64$]{\label{fig:embedding_size_1}
			\includegraphics[width=0.35\columnwidth]{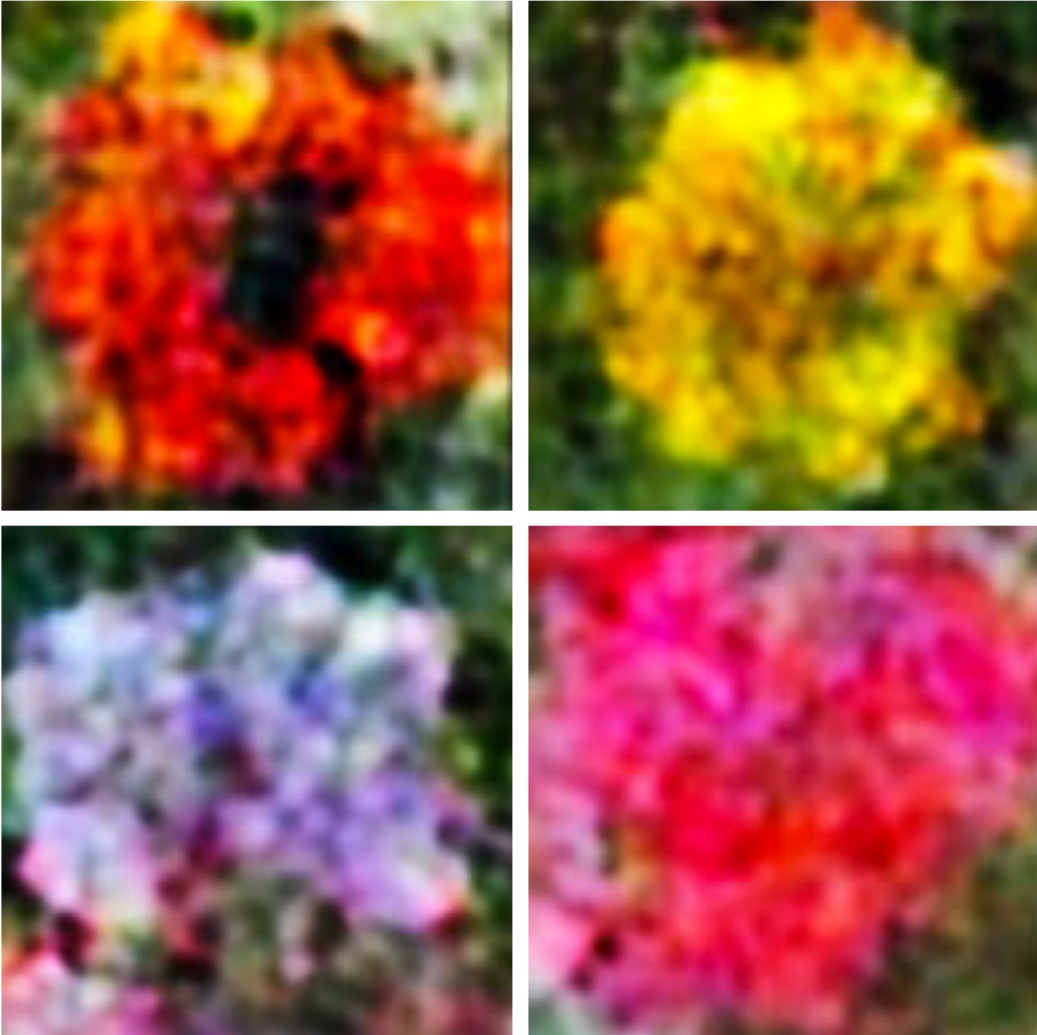}
		}
		\subfigure[$32 \times 32 \times 64$]{\label{fig:embedding_size_2}
			\includegraphics[width=0.35\columnwidth]{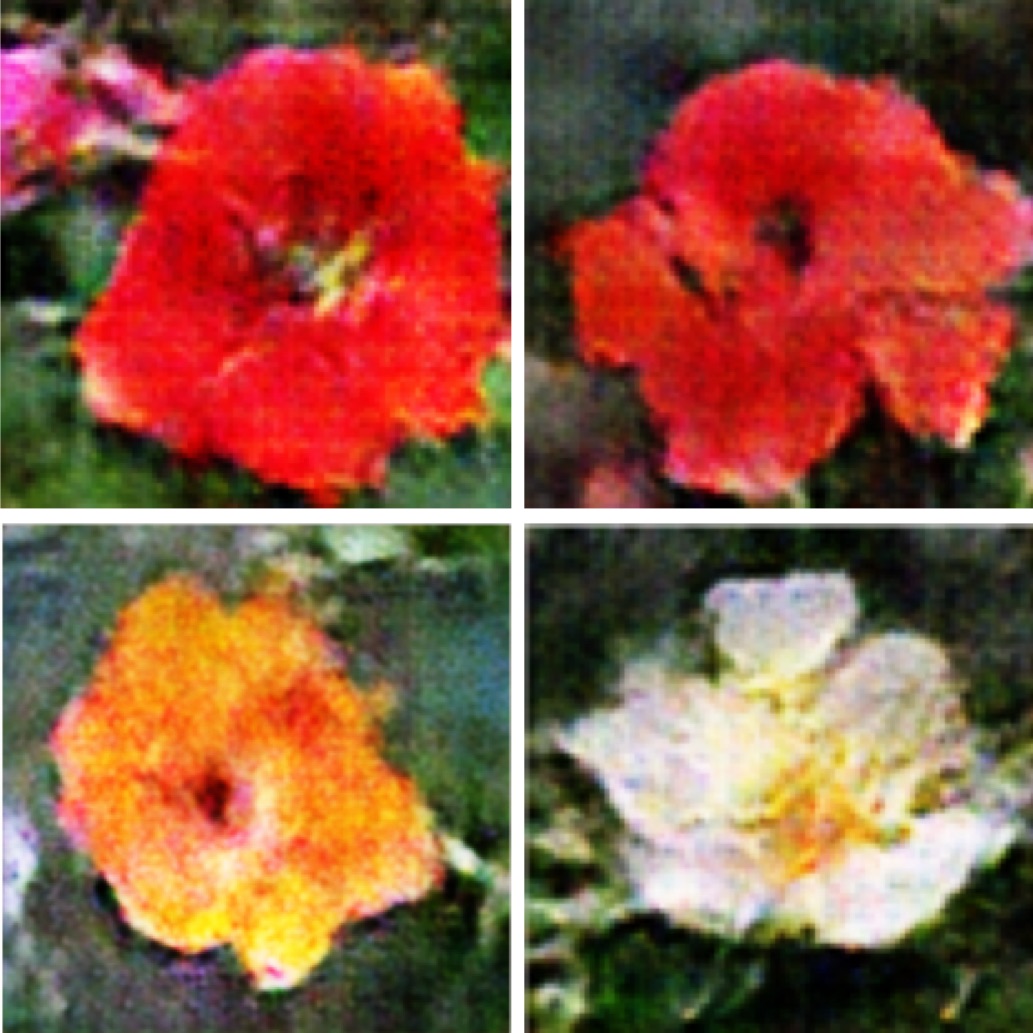}
		}
		\caption{Decoded images from different embedding dimensions.}
		\label{fig:embedding_size}
	\end{figure}

	In this section, we investigate the effect of embedding dimension on the performance of AEGAN.
	Actually, the dimension of the extracted embedding indicates the representation ability of latent space. 
	Although the representation ability will become more powerful with the increase of dimension, the computational complexity will increase accordingly.
	Therefore, a smaller dimension is preferable when the representation ability of the embedding is sufficient. 
	{Here, we {test} smaller embedding dimensions of $8 \times 8$ with 64 channels in the autoencoder.}
	The results are shown in Fig.~\ref{fig:embedding_size}.
	
	Fig.~\ref{fig:embedding_size} {shows that,} when we map the input noise to {such a small embedding}, the final decoded images contain much more visual artifacts than the standard setting in previous experiments, {\it i.e.}, $32 \times 32$ with 64 channels.
	{These results demonstrate that to synthesize a high resolution image, we {must use} a {sufficiently} large {embedding} space to represent the information of style and structure.}
	
	\subsection{Results of Latent Space Interpolation}\label{exp:interpolation}
	{In this section, we conduct an experiment {in which we} investigate the landscape of the latent space. Following the settings in~\cite{radford2015unsupervised}, we conduct linear interpolations between two data points in the latent space and feed them into the generative models.}
	{The generated samples are shown in Fig.~\ref{fig:interpolation}.}
	{In} Fig.~\ref{fig:interpolation}, 
	{there are no sharp transitions and the generated images change smoothly.}
	{These results demonstrate that the proposed AEGAN generalizes well to unseen data}
	rather than simply memorizing the training samples. 
	
	\subsection{Comparison of Training Time}\label{exp:training_time}
\begin{figure}[t]
		\centering
		\includegraphics[width=1.0\columnwidth]{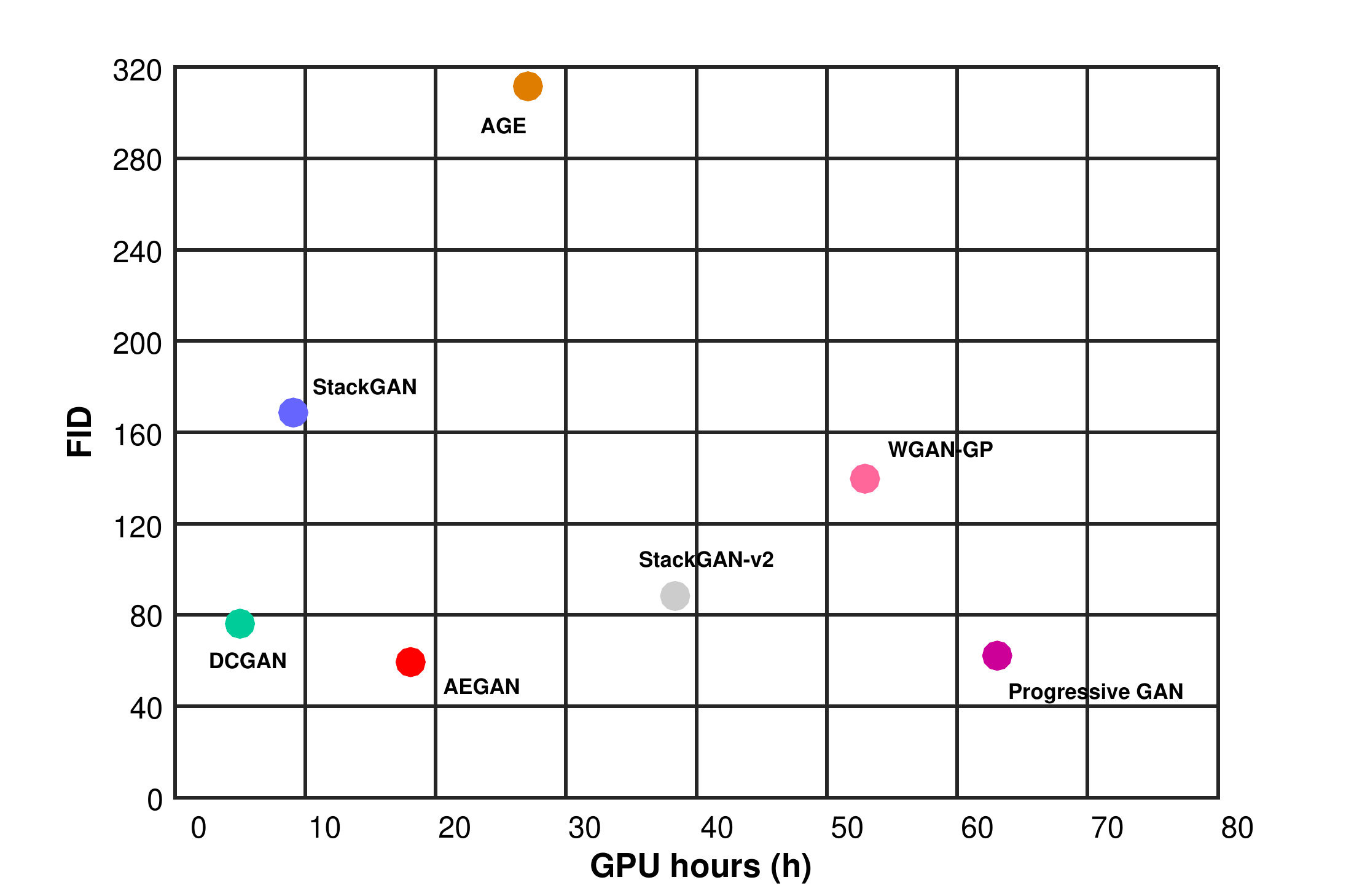}
		\caption{Performance vs training time of different GAN methods on Oxford-102 dataset.}
		\label{fig:training_time}
	\end{figure}

	{
		In this section, we compare the training time of the considered methods. We train all the methods except Progressive GAN on single TITAN X Pascal GPU. For Progressive GAN, we train the model on two GPUs with {a} batch size of 32 {for} each of {GPU} (\textit{i.e.}, 64 in total).
		The training times and the corresponding FID scores are shown in Fig.~\ref{fig:training_time}.
	}
	
	{
		From Fig.~\ref{fig:training_time}, the proposed AEGAN achieves the best FID score of 65.41 and only requires 17 hours {for} training, which is much more efficient than WGAN-GP and Progressive GAN. These results demonstrate the effectiveness of the proposed method in terms of both image generation {performance} and training efficiency.
	}
	
	\section{Conclusion}
	In this paper, we have proposed a novel scheme for high resolution image synthesis. In contrast to traditional GAN methods, 
	{we use low-dimensional embedding to bridge the distribution gap between input noise and real data.}
	Furthermore, we also devise a denoiser network {that} removes the noisy artifacts and provides low-level photo-realistic details in the generated images. The proposed method produces images with 
	{sharp structures and rich photo-realistic details,}
	{and} significantly outperforms the considered generative models.

\end{document}